\documentclass[11pt]{article}

\usepackage[preprint]{acl}

\usepackage{times}
\usepackage{latexsym}
\usepackage[T1]{fontenc}
\usepackage[utf8]{inputenc}
\usepackage{microtype}
\usepackage{inconsolata}
\usepackage{amsmath}
\usepackage{amssymb}
\usepackage{amsthm}
\newtheorem{proposition}{Proposition}
\usepackage{booktabs}
\usepackage{multirow}
\usepackage{graphicx}
\graphicspath{{figures/}}

\makeatletter
\g@addto@macro\UrlBreaks{\do\/\do\-\do\_\do\.}
\makeatother

\title{Differential Backpack Language Models:\\
Sense Inhibition via Signed Contextualization Weights}

\author{Konstantinos Skianis \\
  University of Ioannina \\
  \texttt{kskianis@cse.uoi.gr} \\\And
  Giannis Nikolentzos \\
  University of Peloponnese \\
  \texttt{nikolentzos@uop.gr} \\\AND
  Polykarpos Meladianos \\
  MBZUAI \\
  \texttt{polykarpos.m@mbzuai.ac.ae} \\\And
  Michalis Vazirgiannis \\
  \'Ecole Polytechnique \& MBZUAI \\
  \texttt{michalis.vazirgiannis@mbzuai.ac.ae}}

\begin{document}
\maketitle

\begin{abstract}
Backpack language models predict with a strictly linear combination of
non-contextual per-word sense vectors, weighted by non-negative
contextualization scores. We show that non-negativity conflates two
properties: the linear readout, which carries the interpretability, and the
conic combination, which merely restricts it. We relax the latter by computing the sense weights with differential
attention, a difference of two softmax maps with a learned per-sense gain,
so context can actively inhibit a sense; we call the result a differential
Backpack. At equal parameters on WikiText-103 this cuts the Backpack's
perplexity gap to a Transformer by $31\%$, improves over the softmax
Backpack at every sense count, persists with a shrinking margin up to
$350$M parameters on OpenWebText, and is localized at the sense layer by
ablations and controls. The learned inhibition is
directly interpretable (a financial context drives the river sense of
\textit{bank} to $-0.50$), and the preserved per-sense attribution predicts
counterfactual removals exactly. We tested a number of combination rules, showing the linear interface is nearly free. Making the readout nonlinear gains only $0.43$ perplexity
points while invalidating two
thirds of the decomposition. The differential Backpack's senses are also better
representations, improving zero-shot sentence similarity by up to $33\%$.
% Code link, hidden for now. Restore before submission:
% \footnote{Code:
% \url{https://anonymous.4open.science/r/anon_differential_backpack_lms-E065/}}
\end{abstract}

% Full-width landscape alternative: swap this figure for
%   \begin{figure*}[t] \centering
%   \includegraphics[width=\textwidth]{fig_architecture.png}
%   ...same caption... \end{figure*}
\begin{figure}[t]
  \centering
  \includegraphics[width=\columnwidth]{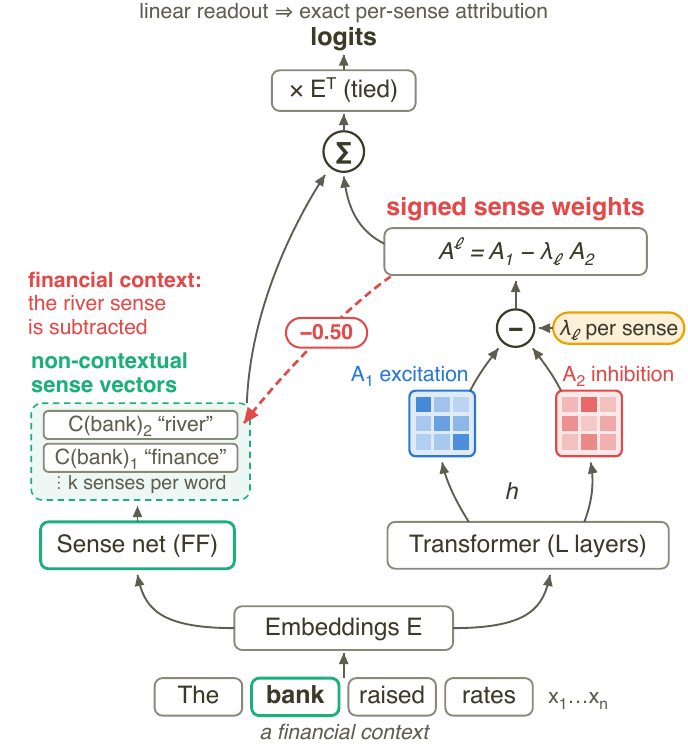}
  \caption{The differential Backpack. Non-contextual sense
  vectors $C(x_j)_\ell$ (left) are combined by signed weights
  $A^\ell = A_1 - \lambda_\ell A_2$ (right) under a linear readout; the
  dashed arrow marks a negative weight suppressing a contextually wrong
  sense.}
  \label{fig:arch}
\end{figure}

\section{Introduction}
\label{sec:intro}

Most interpretability for language models is post hoc. The model is
trained for performance and reverse-engineered afterwards. Backpack language models
\citep{hewitt-etal-2023-backpack} take the opposite route, building an
interpretable interface into the architecture. Each vocabulary item is
assigned $k$ non-contextual sense vectors, and the model's
prediction at every position is a linear combination of the sense vectors
of the preceding tokens (Eq.~\ref{eq:backpack}). Because the readout is
linear in the senses, the contribution of any sense of any word to any
prediction is an exact, inspectable quantity, enabling sense-level
control such as debiasing by down-weighting a targeted sense.

Backpack language models have one notable limitation. The contextualization weights that
combine senses are softmax outputs, hence non-negative, so context can
select a sense or ignore it but never oppose one. When \textit{bank} appears in a
financial sentence, its river sense does not simply go unused. Left
un-suppressed, it leaks into the prediction. A non-negative combination has
no mechanism to cancel this wrong-sense contribution.

This is the setting differential attention was designed for. The DIFF Transformer \citep{ye2024differential}
computes attention as the difference of two softmax maps, so that scores
can be negative and common-mode ``attention noise'' cancels. In a Backpack model, 
noise can be seen as the contribution of the wrong word sense, and the
difference of softmaxes is a mechanism for sense inhibition. We adopt it
without giving up the interface, because non-negativity conflates two
distinct properties: the
linear readout, which carries the interpretability, and the
conic (non-negative) combination, which merely restricts it. We keep the first and relax the second.

We call this differential sense contextualization
(Figure~\ref{fig:arch}). On WikiText-103
at equal parameters it recovers $31\%$ of the perplexity gap
between the softmax Backpack and an unconstrained Transformer, and the
improvement persists, with a shrinking margin, up to $350$M parameters on
OpenWebText. An ablation shows the gain is
specific to the sense-combination layer, exactly where the
non-negativity constraint binds; placing differential attention in the
underlying Transformer instead does not help, and the learned inhibition
is directly interpretable (\S\ref{sec:experiments}).

Our contributions are:
\begin{itemize}
  \item We introduce differential sense contextualization: the Backpack's
    non-negative sense weights become the difference of two softmax maps
    with a learned per-sense gain, so context can inhibit a sense while the
    readout, and hence exact per-sense attribution, stays strictly linear.
  \item At equal parameters the differential Backpack outperforms the softmax
    Backpack on WikiText-103 at every sense count and at $170$M and $350$M
    on OpenWebText. An ablation localizes the gain at the sense layer, and
    three controls that each remove one ingredient recover at most $28\%$
    of it and an existing signed rule none, so the gain is caused by the
    differential form of inhibition.
  \item We make the expressivity of negative attention
    \citep{understanding-diff} precise for sense combination,
    characterizing which outputs each rule can reach and showing the
    constraint binds at the sense layer but not in a Transformer, with a
    separation that is exact at one sense head and decays with more, which
    we confirm.
  \item We examine a number of combination rules that decompose the Backpack's perplexity
    tax. Only $18\%$ is attributable to the linear interface, and the rest
    is depth that the sense dictionary's parameters displace; at matched
    depth the differential Backpack trails a Transformer by $0.14$
    perplexity points. Differential
    contextualization is the best exactly-attributable rule we test, its
    attribution predicts counterfactual removals exactly where the
    nonlinear rung's does not, and its sense outputs are the strongest
    sentence representations we measure at $70$M and $170$M.
  \item The learned inhibition is analyzable and usable. The gain
    $\lambda$ reaches the same value from three initializations, is
    highest when senses are few, and matters above $1$ ($0.32$
    perplexity points at $k{=}4$); suppressing a targeted sense at inference cuts gender bias
    by up to $39.7\%$ at negligible cost, most effectively at scale. A
    zero-shot WiC probe shows no advantage, which bounds the claim.
\end{itemize}

\section{Related Work}
\label{sec:related}

\paragraph{Backpack language models.}
Backpacks \citep{hewitt-etal-2023-backpack} were introduced as an
architecture coupling Transformer-level performance with a linear,
sense-level interface, and demonstrate sense-based control including
targeted debiasing. Follow-up work extends the architecture to
character-level Chinese \citep{sun2023chinese}, uses sense finetuning for
model editing from canonical examples \citep{hewitt2024model}, adapts
senses across languages \citep{cruz2026multilingual}, and intervenes on
senses to debias retrieval \citep{afzali2026backpack}. All of this work keeps the non-negative contextualization weights; we
modify that mechanism itself, narrowing the perplexity gap to Transformers
while preserving the sense interface these methods rely on. A complementary line
asks whether the interface must be trained in at all, inducing sense
pathways into frozen pretrained models \citep{cruz2026sense}; we find that senses read off a trained Transformer's output-value (OV) circuits
\citep{elhage2021framework} are weaker than learned ones on every lexical
benchmark we test.

\paragraph{Differential attention.}
The Differential Transformer (DIFF Transformer, \citealp{ye2024differential}) computes attention
as a difference of two softmax maps to cancel common-mode noise, and has
prompted gated \citep{diff-gated}, parameter-shared \citep{shared-diff}
and thresholded \citep{huang2026threshold} variants. Cog Attention
\citep{lv2024cog} obtains negative weights by a different rule,
$\mathrm{sign}(qk)$ times a softmax over $|qk|$, so a row's absolute weights sum to one (we test it at the sense layer in
\S\ref{sec:controls}); the Integral Transformer
\citep{kobyzev-etal-2025-integral} argues that both routes risk discarding
useful tokens. A subsequent study finds the DIFF Transformer's
advantage stems from negative-attention expressivity and reduced head
redundancy rather than denoising per se, with modest gains at small scale
\citep{understanding-diff}. Our results agree that differential attention
helps at the sense layer, not in the underlying Transformer. None of this
work places signed attention
at a linear, inspectable readout, where its expressivity can be characterized exactly (\S\ref{sec:theory}).

\paragraph{Word senses as explicit vectors.}
Representing a word by several vectors instead of one has a long history:
multi-prototype and non-parametric sense embeddings cluster a word's
contexts and learn a vector per cluster
\citep{reisinger-mooney-2010-multi,neelakantan-etal-2014-efficient}, a
literature surveyed by \citet{camacho-collados-pilehvar-2018-survey}. Those
senses live outside the language model and are selected one at a time by a
similarity rule, whereas a Backpack makes them components of the model's own prediction. 
Unsupervised sense induction still fails to beat a one-cluster-per-lemma
baseline even with modern language models \citep{mosolova2026wsi}, so an
architecture that carries an explicit inventory is still a live design
choice rather than a legacy of pre-neural lexical semantics. Differential contextualization adds
the operation that selection-based schemes structurally lack: a sense can be scored below zero, not merely passed over.

\paragraph{Interpretability by architecture.}
Our work sits with approaches that make a model interpretable by
construction, not by post-hoc probing. The contrast is sharpest
against attention weights, contested as explanations because the map from
weights to outputs is nonlinear and non-unique
\citep{jain-wallace-2019-attention,wiegreffe-pinter-2019-attention}.
Backpack sense weights are not proxies in this sense. They multiply fixed
vectors under a linear readout, so a weight is a contribution. A
second line accepts the trained network as given and recovers an
interpretable basis after the fact, from superposition analyses
\citep{elhage2022superposition} to sparse dictionary learning over
activations \citep{huben2024sparse}. Our ladder bears on that choice. The cost of a Backpack is its fixed
per-token dictionary, not its linear readout.

\section{Method}
\label{sec:method}

\subsection{Background: Backpack Language Models}
Given a sequence $x_{1:n}$, a Backpack \citep{hewitt-etal-2023-backpack}
represents each vocabulary item $x$ with $k$ non-contextual sense vectors
$C(x) \in \mathbb{R}^{k \times d}$ and predicts from a linear combination
\begin{equation}
  o_i = \sum_{\ell=1}^{k} \sum_{j \le i} A^{\ell}_{ij}\, C(x_j)_\ell,
  \qquad
  \mathrm{logits}_i = o_i E^\top,
  \label{eq:backpack}
\end{equation}
where $E$ is the (tied) embedding matrix and the contextualization
weights $A^{\ell} \in \mathbb{R}^{n \times n}$ are computed by an
attention-style map over the hidden states $h$ of an underlying
Transformer \citep{vaswani2017attention}, one head per sense. In the original model each row of
$A^{\ell}$ is a softmax output: $A_{ij}^{\ell} \ge 0, \, \forall i,j \in \{ 1, \dots, n\}$.

\subsection{Our proposed approach: Differential Sense Contextualization}
We compute the sense weights as a difference of two causal softmax maps
(Figure~\ref{fig:arch}):
\begin{equation}
  A^{\ell} = \mathrm{softmax}\!\left(\tfrac{Q_1 K_1^\top}{\sqrt{d_h}}\right)
  - \lambda_\ell\, \mathrm{softmax}\!\left(\tfrac{Q_2 K_2^\top}{\sqrt{d_h}}\right),
  \label{eq:diff}
\end{equation}
with a learned per-sense gain reparameterized as
$\lambda_\ell = \exp(\lambda_{q_1}\!\cdot\!\lambda_{k_1})
- \exp(\lambda_{q_2}\!\cdot\!\lambda_{k_2}) + \lambda_{\mathrm{init}}$
following \citet{ye2024differential}. Weights become signed. The first
map excites senses, the second inhibits them; each entry
lies in $[-\lambda_\ell, 1]$ and each row's mass is $1 - \lambda_\ell$.

\paragraph{Preserving the linear readout.}
The signed weights leave Eq.~\ref{eq:backpack} otherwise intact, so
$\mathrm{logits}_i$ is still an exact linear function of the sense vectors:
the contribution of sense $\ell$ of token $j$ to the prediction at $i$ is
exactly $A^{\ell}_{ij}\,(C(x_j)_\ell E^\top)$, now carrying a sign. We
apply no normalization after the sense combination, keeping the per-head
RMSNorm of the DIFF Transformer inside the underlying Transformer
only, since any normalization of $o_i$ would make the readout nonlinear in
$C$; sense vectors are LayerNormed at creation instead. The second map
costs a second query/key projection at the sense layer, $0.5$M parameters
($0.8\%$ of the model at $70$M) and $6\%$ of forward FLOPs, and
$12$--$16\%$ of training throughput in practice
(Appendix~\ref{app:hparams}).

\paragraph{Terminology.}
Differential attention is the mechanism of Eq.~\eqref{eq:diff}, a
difference of two softmax maps. A DIFF Transformer \citep{ye2024differential} applies it inside a
Transformer's self-attention layers; a differential Backpack, our model,
applies it only to the sense contextualization weights and keeps softmax
attention in its underlying Transformer, while the softmax Backpack is the
original.
In tables and figures, softmax and diff name a Backpack's sense-weight
rule, never a Transformer's attention.

\paragraph{Why the constraint binds only in Backpacks.}
Non-negativity is a weak constraint in a standard Transformer. Signed
value and output projections let a head realize subtraction downstream, so
negative attention adds little the architecture cannot already route
around. A Backpack is different. Its sense-combination layer is the single
point where context integrates lexical information, over frozen,
non-contextual sense vectors the model cannot reshape, so non-negativity
binds (there is no way to express ``not this sense'') and relaxing it adds expressivity that has no substitute. This
predicts that differential attention should help at the sense layer but not
in the underlying Transformer, which our ablation confirms
(\S\ref{sec:experiments}).

\subsection{An Expressivity Separation}
\label{sec:theory}
A softmax Backpack averages the sense vectors in context and cannot leave
the region they span; signed weights can, since $1.5\,u - 0.5\,v$ sits past
$u$. A non-negative model can say ``mostly finance, partly river''; only a
signed one can say ``finance, and not river''.

That signed attention is more expressive has been observed empirically
\citep{understanding-diff}; what follows makes it exact for sense
combination and identifies where it binds. Formally, fix a position $i$ and
write $S_\ell = \{C(x_j)_\ell : j \le i\}$ for the sense-$\ell$ vectors
available in context.

\begin{proposition}[Reachable sets]
\label{prop:reach}
For the position $i$ fixed above and a sense $\ell$, let
$b = (A^{\ell}_{i1}, \dots, A^{\ell}_{ii}) \in \mathbb{R}^i$ be the row of
weights that position $i$ assigns to the $i$ available positions, and let
$\Delta = \{p \in \mathbb{R}^i : p \ge 0,\ \sum_j p_j = 1\}$ be the
probability simplex. For a softmax Backpack $b \in \Delta$ for every
$\ell$, so $o_i \in \sum_{\ell=1}^{k}\mathrm{conv}(S_\ell)$, a Minkowski
sum of convex hulls. For a differential Backpack with gain
$\lambda_\ell \in (0,1)$ and $i \ge 2$, a vector $b$ is realizable (in
closure) as $p - \lambda_\ell q$ with $p,q \in \Delta$ if and only if
\begin{equation}
  \textstyle\sum_j b_j = 1-\lambda_\ell
  \quad\text{and}\quad
  \sum_j \max(-b_j, 0) \;\le\; \lambda_\ell .
  \label{eq:reach}
\end{equation}
Equivalently, the normalized coefficients $a = b/(1-\lambda_\ell)$ sum to
one and have negative part $\sum_j \max(-a_j,0) \le
\lambda_\ell/(1-\lambda_\ell)$; since $\|a\|_1 = 1 + 2\sum_j\max(-a_j,0)$
on that hyperplane, they range over the $\ell_1$ ball section
\begin{equation}
  \mathcal{A}_{\lambda_\ell} =
  \Big\{a : \textstyle\sum_j a_j = 1,\;
  \|a\|_1 \le \tfrac{1+\lambda_\ell}{1-\lambda_\ell}\Big\}
  \;\supsetneq\; \Delta ,
  \label{eq:l1ball}
\end{equation}
with the bound attained at $p = e_1$, $q = e_2$.
\end{proposition}

\noindent The proof is in Appendix~\ref{app:proofs}. Relaxing
non-negativity trades the
simplex for an $\ell_1$ ball of radius set by $\lambda_\ell$, the geometry
of signed sparse coding: the softmax Backpack can only interpolate among the
senses present in context, whereas signed weights can extrapolate past
them.

\begin{proposition}[Strict separation at $k{=}1$]
\label{prop:sep}
Let $k{=}1$, let $u,v$ be orthonormal, and let $\lambda \in (0,1)$. Suppose
the data pin the single-token outputs $o$ to $u$ and $v$ respectively (a
single-token context fixes the output exactly, since only one weight is
available), and require the two-token output $o = (1+s)u - sv$ for some
$s>0$. Every softmax Backpack incurs squared Euclidean error
$\|o - ((1+s)u - sv)\|^2 \ge 2s^2$ (a distance of $s\sqrt{2}$), while a
differential Backpack attains $0$ if and only if $\lambda \ge s/(1+s)$.
\end{proposition}

\noindent The proof is in Appendix~\ref{app:proofs}. The separation is governed by $\lambda$, which the model grows during
training (\S\ref{sec:lambda}). Both propositions assume
$\lambda_\ell < 1$, so that the row sum is positive. Trained models
sometimes exceed it (mean $\lambda$ above $1$ at $k{=}4$, individual senses
reaching $2.15$); such a sense is net inhibitory, subtracting a weighted
average of the context rather than adding one, and the separation argument
no longer applies (Appendix~\ref{sec:lambda}). The regime is not
incidental. Constraining $\lambda$ to $(0,1)$ costs $0.32$ perplexity points at $k{=}4$
but nothing at $k{=}16$; characterizing it remains open. The bound is tight
only at $k{=}1$: with more heads a softmax Backpack can imitate signed
behaviour by cancelling heads, since single-token contexts pin only
$\sum_\ell C(x)_\ell$. We expect a separation exact at $k{=}1$ that decays
with $k$, and measure it in \S\ref{sec:synthetic}.

\section{Experiments}
\label{sec:experiments}

\subsection{Experimental Setup}
All models are trained from scratch on WikiText-103
\citep{merity2017pointer} with the GPT-2 byte-pair encoding (BPE) vocabulary
\citep{radford2019language} and a context of $512$ tokens, unless stated
otherwise. The main configuration has $\sim$70M parameters ($d{=}512$, an
$8$-layer underlying Transformer, $k{=}16$ senses) and is trained for
$20$k steps with AdamW, a cosine schedule and bf16 precision; Transformer
baselines use a $14$-layer trunk to match parameters ($70.1$M vs.\
$69.6$M). Larger models are trained on OpenWebText
\citep{gokaslan2019openwebtext} at $170$M and $350$M parameters
(Appendix~\ref{sec:owt}). We report exact validation perplexity (ppl)
over non-overlapping windows of the full validation split.

Running the same evaluation battery on the released Backpack-GPT2
\citep{hewitt-etal-2023-backpack} reproduces its published lexical profile
and the properties our analysis assumes (Appendix~\ref{sec:spec}).

\subsection{Main Result: Equal-Parameter Comparison}
\begin{table}[t]
\centering
\small
\setlength{\tabcolsep}{4pt}
\begin{tabular}{lccc}
\toprule
Model & Params & PPL $\downarrow$ & tax \\
\midrule
Transformer                & 70.1M & 21.09 & --- \\
DIFF Transformer           & 70.1M & 21.00 & $-0.09$ \\
softmax Backpack            & 69.6M & 24.60 & $+3.51$ \\
differential Backpack (ours) & 69.6M & \textbf{23.52} & $+2.43$ \\
\quad + DIFF trunk          & 69.6M & 24.78 & $+3.69$ \\
\bottomrule
\end{tabular}
\caption{WikiText-103 exact validation perplexity at matched
parameters and identical training budget (20k steps, single seed); tax is
the gap to the plain Transformer.}
\label{tab:main}
\end{table}

\begin{figure}[t]
  \centering
  \includegraphics[width=\columnwidth]{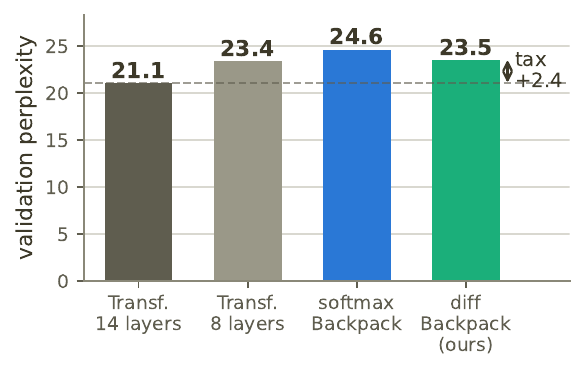}
  \caption{Interpretability tax on WikiText-103: exact validation
  perplexity at matched parameters ($\sim$70M), with a Transformer of the
  Backpack trunk's depth ($8$ layers, $51$M) for reference.}
  \label{fig:tax}
\end{figure}

Differential sense contextualization (Table~\ref{tab:main},
Figure~\ref{fig:tax}) improves exact validation perplexity
from $24.60$ to $23.52$, cutting the Backpack's gap to the plain Transformer from $3.51$ to
$2.43$, a $31\%$ reduction at equal parameters and budget.
The controls isolate the mechanism. A DIFF Transformer matches the plain
Transformer ($21.00$ vs.\ $21.09$), and adding differential attention to
the differential Backpack's underlying Transformer (``+ DIFF trunk'') is
worse than the softmax Backpack, so the improvement comes from signing
the sense weights. The result persists on OpenWebText: $23.15 \pm 0.03$ vs.\ $23.84 \pm
0.13$ over three seeds at $170$M, and $19.96$ vs.\ $20.48$ at $350$M, where
a parameter-matched $24$-layer Transformer reaches $18.02$, so the tax falls
from $2.46$ to $1.94$ ($21\%$). The relative gain shrinks with scale
($4.4\%$, $2.9\%$, $2.5\%$) while inhibition usage does not
(Appendix~\ref{sec:owt}).

Retraining both headline arms with three seeds gives $24.60 \pm 0.12$ and
$23.52 \pm 0.04$ (mean $\pm$ s.d.); the $1.08$ ppl gap is $14$ standard
errors (Welch's $t{=}14.4$, $p{<}0.005$).

\subsection{Decomposing the Interpretability Tax}
\label{sec:ladder}
We locate the source of the Backpack's gap by placing the model on a ladder
of combination rules sharing the main configuration and differing only in
how sense weights are formed. Five rungs keep predictions linear in the senses, so attribution stays
exact: hard selection (one-hot, straight-through), non-negative (softmax),
signed by difference (diff), signed by sign and magnitude (Cog Attention,
\citealp{lv2024cog}), and unconstrained weights (raw real-valued scores). A sixth applies a residual multilayer perceptron (MLP) to the sense sum, the one change that makes
predictions nonlinear in the senses.

\begin{table}[t]
\centering
\small
\setlength{\tabcolsep}{4pt}
\begin{tabular}{lccc}
\toprule
Combination rule & PPL $\downarrow$ & tax & attr.\ resid. \\
\midrule
Transformer            & 21.09 & ---    & n/a \\
DIFF Transformer       & 21.00 & $-0.09$ & n/a \\
Transformer, 8 layers  & 23.38 & $+2.29$ & n/a \\
\midrule
one-hot                & 31.58 & $+10.49$ & $2\!\cdot\!10^{-7}$ \\
softmax                & 24.60\,$\pm$.12 & $+3.51$ & $5\!\cdot\!10^{-7}$ \\
\textbf{diff Backpack (ours)} & \textbf{23.52\,$\pm$.04} & $+2.43$ & $6\!\cdot\!10^{-7}$ \\
unconstrained weights  & 26.67 & $+5.58$  & $2\!\cdot\!10^{-6}$ \\
Cog Attention          & 24.72 & $+3.63$  & $4\!\cdot\!10^{-7}$ \\
\midrule
softmax\,+\,MLP        & 23.97\,$\pm$.17 & $+2.88$ & $0.67$ \\
diff\,+\,MLP           & 23.09\,$\pm$.08 & $+2.00$ & $0.66$ \\
\bottomrule
\end{tabular}
\caption{The constraint ladder on WikiText-103. All rows share the main
configuration and budget; tax is the gap to the $14$-layer Transformer;
the $8$-layer Transformer ($51$M) matches the Backpack trunk's depth; attr.\
resid.\ is the relative error of the per-sense decomposition against the
model's own logits.}
\label{tab:ladder}
\end{table}

Table~\ref{tab:ladder} yields three findings, the most important being
that linearity is cheap. Making the readout nonlinear improves the
differential Backpack from $23.52$ to $23.09$, a gain of $0.43$ ppl
($7.8$ standard errors over three seeds), while the
attribution residual jumps from $6\!\times\!10^{-7}$ to $0.66$. The
remaining $2.00$ ppl of tax survives the MLP, so only $18\%$ of the gap to a
Transformer is attributable to the linear interface; the other $82\%$ is
not. That remainder is depth. A Transformer with the Backpack trunk's
$8$ layers ($51$M parameters) reaches $23.38$, so the six layers alone cost
$2.29$ ppl, and against it the differential Backpack trails by only
$0.14$ ppl while carrying $19$M more parameters in its non-contextual
dictionary; with the MLP it leads by $0.29$. The tax at matched parameters
is therefore the depth that the dictionary's parameters displace, not the
linear interface.

Constraints also help optimization. Removing them entirely is worse than
softmax ($26.67$ vs.\ $24.60$) and hard selection is far worse ($31.58$), both rungs trained to the same
budget. The differential relaxation
loosens the simplex sufficiently to permit inhibition while retaining
normalization.

Among the exactly-attributable rules, differential contextualization is
the strongest.

\subsection{Controls for the Causal Claim}
\label{sec:controls}
Differential contextualization changes three things at once, two
attention maps, a learned gain and signed weights; only the third is
inhibition, so three controls each remove one while matching the rest. The sum control adds the two
maps, $A_1 + \mathrm{softplus}(\lambda_\ell)A_2$, keeping the two-map
structure and parameter count but leaving weights non-negative.
The gain control scales a single softmax by a learned positive per-sense factor,
relaxing the row-sum constraint without permitting signs. The $\lambda{=}0$
control freezes the differential gain at zero, leaving the second map
without gradient: the differential parameterization with inhibition
switched off, which doubles as a noise check. A fourth rule, Cog Attention \citep{lv2024cog}, is signed by another
route: $\mathrm{sign}(qk)$ times a softmax over $|qk|$, a single map whose
absolute row sum is one.

\begin{table}[t]
\centering
\small
\setlength{\tabcolsep}{4pt}
\begin{tabular}{lccc}
\toprule
Sense weights & PPL $\downarrow$ & \% of gain & neg.\ mass \\
\midrule
softmax                 & 24.60 & ---   & 0.00 \\
\midrule
two maps, added         & 24.30 & 28\%  & 0.00 \\
scaled softmax          & 24.51 & 9\%   & 0.00 \\
$\lambda{=}0$           & 24.55 & 5\%   & 0.00 \\
Cog Attention           & 24.72 & $-11\%$ & 0.39 \\
\midrule
\textbf{diff Backpack (ours)} & \textbf{23.52} & \textbf{100\%} & 0.45 \\
\bottomrule
\end{tabular}
\caption{Controls for the causal claim (WikiText-103, main
configuration). ``\% of gain'' is the share of diff's $1.08$ ppl
improvement over softmax each rule recovers.}
\label{tab:controls}
\end{table}

Table~\ref{tab:controls} isolates the cause. The two-map structure alone
recovers $28\%$ of the gain and the relaxed row sum $9\%$, while the
$\lambda{=}0$ run lands $0.05$ ppl from softmax, within the $0.12$ seed
deviation; all three have zero negative weight mass, so each removed
inhibition and nothing else. Roughly three quarters of the
improvement is therefore due to signed weights themselves. Signedness alone does not suffice: Cog Attention lands at $24.72$,
softmax level, despite $0.39$ negative weight mass; what recovers the gain
is the difference-of-softmax structure with its positive row sum, the
geometry of Proposition~\ref{prop:reach}. The controls are single-seed; the smallest gap they must explain ($0.30$
ppl) far exceeds the $0.12$ seed deviation.

\subsection{Downstream Probes}
\label{sec:downstream}
Perplexity need not track usefulness, so we probe frozen representations
on five tasks: part-of-speech (POS) tagging (UD English-EWT
\citep{silveira-etal-2014-gold}, linear probe), STS-B sentence similarity
\citep{cer-etal-2017-semeval} (zero-shot cosine of mean-pooled
representations, no trained parameters), SST-2 \citep{socher-etal-2013-recursive} and MRPC
\citep{dolan-brockett-2005-automatically} (linear probes, GLUE validation splits), and clustering contextual
occurrences of ambiguous words against WiC sense labels (adjusted Rand
index). Each is run on the underlying Transformer state $h$
and on the sense output $o$, the Backpack's actual interface.

\begin{table}[t]
\centering
\small
\setlength{\tabcolsep}{4pt}
\begin{tabular}{lcccc}
\toprule
Representation (70M) & POS & STS & SST-2 & ARI \\
\midrule
Transformer ($h$)       & 0.859 & 0.488 & 0.753 & 0.174 \\
DIFF Transformer ($h$)  & 0.859 & 0.499 & 0.743 & 0.203 \\
\midrule
softmax Backpack ($o$)   & 0.844 & 0.372 & 0.718 & 0.151 \\
diff Backpack ($o$)      & 0.815 & 0.495 & 0.731 & 0.174 \\
Cog Backpack ($o$)       & 0.836 & 0.422 & 0.736 & 0.163 \\
\midrule
diff Backpack + MLP ($o$) & 0.838 & 0.485 & 0.749 & 0.168 \\
\bottomrule
\end{tabular}
\caption{Frozen-representation probes at $70$M: POS accuracy (UD
English-EWT), STS-B Spearman, SST-2 probe accuracy, and sense-clustering
ARI, on the Transformer state $h$ or the sense output $o$; all scales and
MRPC in Appendix~\ref{sec:owt}.}
\label{tab:downstream}
\end{table}

Table~\ref{tab:downstream} reports the $70$M probes. The semantic result
is the strongest. The differential Backpack's sense outputs reach $0.495$
Spearman on STS-B against $0.372$ for the softmax Backpack's ($+33\%$
relative), $0.554$
against $0.466$ at $170$M ($+19\%$), and $0.539$ against $0.476$ at
$350$M ($+13\%$). Against the Transformers' hidden states, which carry no
attribution interface, the differential Backpack's interface is within
noise on STS-B ($0.495$ vs.\ $0.488$ and $0.499$; s.e.\ $0.03$) and SST-2,
and trails on POS and clustering, the representational side of the
interpretability tax, which the softmax Backpack incurs in full ($0.372$
on STS-B). At $350$M the Transformer's state ($0.572$) is the stronger
sentence representation.
Differential attention inside a plain Transformer leaves its hidden state unchanged on every probe (STS $0.499$
vs.\ $0.488$), so the representational gain, like the perplexity gain, is
specific to the sense layer; and Cog Attention's signed rule recovers part
of it ($0.422$) but stays well short of diff. Sense clustering agrees at
every scale ($0.174$ vs.\ $0.151$ at $70$M), and SST-2 probes favour the differential Backpack's senses at every scale ($0.731$ vs.\ $0.718$, $0.792$ vs.\
$0.773$, $0.792$ vs.\ $0.786$), each within one standard error ($1.5$
points); MRPC probes stay at the majority class and do not separate models. The trade is syntactic. The differential Backpack's senses are slightly worse for POS tagging ($0.815$ vs.\ $0.844$). The nonlinear rung
improves no probe beyond noise (STS $0.485 < 0.495$, SST-2 $0.749$ vs.\
$0.731$).

\subsection{Sense Quality and Disambiguation}
A sense lens lists the vocabulary items a sense vector most strongly
predicts through the tied embedding, $C(w)_\ell E^\top$, with no probe and
no labelled data. For \textit{rock}, the
differential Backpack devotes one sense to the mineral reading (\textit{rock,
Rocks, canyon, Canyon}) and another to the music-genre reading
(\textit{rock, garage, punk, gig, pop}), whereas the two most
\textit{rock}-specific senses of the softmax Backpack both blend geological
(\textit{shale, bedrock}) and musical (\textit{punk, metal, Stones})
tokens. Appendix~\ref{sec:spec} quantifies this. Senses tracking human
similarity judgments rise from $8/64$ to $37/64$ at $k{=}64$.

\begin{figure}[t]
  \centering
  \includegraphics[width=0.96\columnwidth]{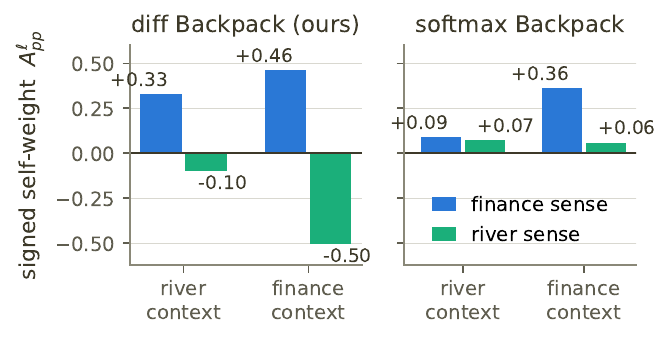}
  \caption{Signed self-weight $A^{\ell}_{pp}$ of the finance
  and river senses of \textit{bank} in a river vs.\ a financial context,
  for the differential (left) and softmax (right) Backpacks.}
  \label{fig:disambig}
\end{figure}

\paragraph{Inhibition can perform word-sense disambiguation.}
For \textit{bank}, take each model's most recruited finance sense (lens
\textit{subsidiary, mint, Treasury} in ours) and its river sense
(\textit{River, river, rivers, stars}).
Figure~\ref{fig:disambig} shows the signed self-weight $A^{\ell}_{pp}$ the
two senses receive as \textit{bank} appears in two disambiguating contexts.
In the differential Backpack they move in opposite directions. A financial
context raises the finance sense from $+0.33$ to $+0.46$ and drives the
river sense from $-0.10$ to $-0.50$, cancelling the contextually wrong
sense. The softmax Backpack, floored at zero, can only move one way,
boosting its finance sense ($+0.09\to+0.36$) with no way to suppress the
river sense ($+0.07\to+0.06$). Signed weights
thus supply a mechanism the non-negative variant lacks: cancellation of
the wrong word sense, differential denoising with a lexical meaning. The
signal is cleanest for words like
\textit{bank} whose senses separate sharply, and \S\ref{sec:inhibit} shows
this is not inhibition's dataset-level role.

\subsection{Faithfulness of the Attribution}
\label{sec:faithful}
At $300$ validation positions we
remove the ten (sense, position) pairs with the largest attributed
contribution to the target logit, one at a time, and compare attributed
with actual change (Table~\ref{tab:faithful}, Appendix~\ref{app:inhibit}).
On every linear rung the two agree to numerical precision; with the MLP the
rank correlation falls to $0.77$--$0.82$ and the median contribution is
mis-sized by $43$--$48\%$, which is what the $0.66$ residual of
Table~\ref{tab:ladder} means behaviourally. It is also useful in the ERASER sense. Removing the ten top pairs costs $3.7$ nats of target
log-probability against $0.01$ for ten random pairs, and keeping only them
retains the prediction to within $1.2$ nats against $7.8$.

\subsection{Inhibition: Structure and Ablation}
\label{sec:inhibit}
\paragraph{Inhibition structure.}
Table~\ref{tab:sense} quantifies how the weights are used. Differential
contextualization places $45\%$ of the weight mass on negative
(inhibitory) weights (zero for softmax by construction) and engages
more of the sense$\times$position grid, not less: $|A|$ entropy is
higher and sparsity lower. This argues against the hypothesis that subtraction sharpens selection:
the gain comes from added expressivity, consistent with
\citet{understanding-diff}. Inhibition is not localized to a subset of senses. Every sense takes both
signs depending on context, with per-sense negative-mass fractions of $0.26$--$0.61$
(Fig.~\ref{fig:neghist}).

Lexical quality is preserved (Table~\ref{tab:sense}): best-sense $\rho$ on
SimLex-999/WordSim-353 is $0.44$/$0.73$ for diff against $0.42$/$0.70$ for
softmax, both far above the raw-embedding baselines.

\paragraph{Is inhibition load-bearing, and where?}
We remove only the subtraction from a trained differential Backpack at
inference, clamping negative weights to zero and rescaling each row to its
original mass, or dropping the inhibition map, and bucket the per-token
loss increase by the WordNet polysemy of the preceding word
(Figure~\ref{fig:inhibit}). The trained model cannot function without
inhibition. Perplexity rises from $23.5$ to $350$ at $70$M, $23.2$ to
$252$ at $170$M and $20.0$ to $376$ at $350$M (the softmax control is
untouched), whereas a model trained without inhibition is only $1$ ppl
worse (Table~\ref{tab:controls}); the senses and the excitation map are
co-adapted to the subtraction. The damage is not where a disambiguation
account predicts. It is largest after non-lexical tokens and decreases
mildly with polysemy ($2.45$ nats after monosemous words, $2.24$ after
words with eight or more senses at $70$M, the same profile at $170$M and
$350$M). Inhibition is a general-purpose part of the combination, of which \textit{bank} is an instance rather than the rule
(cf.\ Appendix~\ref{sec:wic}).

\subsection{Debiasing by Sense Intervention}
\label{sec:control}
Following the sense-intervention protocol of
\citet{hewitt-etal-2023-backpack}, we take $20$ profession nouns, identify
each word's most gendered sense exactly via the linear readout
($\arg\max_\ell |C(w)_\ell E^\top_{\textit{he}} - C(w)_\ell
E^\top_{\textit{she}}|$), and edit that sense's contextualization weights at
inference. We compare three interventions: down-weighting
($A^\ell \!\to\! 0.5A^\ell$), removal ($A^\ell \!\to\! 0$), and
suppression ($A^\ell \!\to\! -|A^\ell|$), the signed operation
that only a differential Backpack performs natively. We report bias as the
mean $|\log p(\textit{he}) - \log p(\textit{she})|$ over six templates.

Since an intervention can always reduce bias by damaging the model, we pair bias with a collateral damage measure, the Kullback--Leibler (KL)
divergence between
edited and clean next-token distributions over the non-gendered
vocabulary, renormalized.

Suppression is the strongest intervention in every model
(Table~\ref{tab:control}), and all edits are surgical. Validation perplexity moves by less than
$0.004$ ppl, so they touch only the targeted senses. The architecture-specific result appears at $170$M and $350$M
(Figure~\ref{fig:control}), where the differential Backpack dominates the
frontier. At $170$M, suppressing a sense reduces bias by $39.7\%$ (0.763 in
absolute terms) at a collateral KL of $0.056$, whereas the softmax Backpack
reaches only $31.4\%$ (0.545) at $0.173$: $1.40\times$ the absolute
effect for less than a third of the damage. At matched damage, softmax removal achieves a $0.535$ reduction while
differential suppression achieves $0.763$, $43\%$ more. Suppression also adds substantially
over removal for the differential Backpack ($27.9\%\!\to\!39.7\%$) but almost
nothing for softmax ($30.8\%\!\to\!31.4\%$), consistent with the inhibition
pathway being one that only a model trained with signed weights exploits.
At $350$M the pattern repeats: $32.5\%$ at KL $0.042$ against $27.7\%$ at
$0.233$ for softmax, and suppression again adds over removal for the
differential Backpack ($22.2\%\!\to\!32.5\%$) but little for softmax
($25.3\%\!\to\!27.7\%$).

Differential Backpacks start from a higher baseline bias, so their advantage
is controllability, not lower bias to begin with. The frontier ordering also reverses at $70$M, where the softmax Backpack is
the more efficient one; our single-seed evidence at each scale therefore
supports the claim from $170$M upward but not as a scale-free property.

\section{Conclusion}
\label{sec:conclusion}
We keep the Backpack's linear readout, which carries its
interpretability, and relax the conic combination that merely restricts
it, computing sense weights as a difference of two softmax maps so that
context can inhibit a sense. At equal parameters this cuts the gap
to an unconstrained Transformer by $31\%$ on WikiText-103 and persists to
$350$M on OpenWebText with a margin that shrinks with scale, with an
ablation localizing the gain at the sense-combination layer. Because the
readout stays linear, the attribution predicts counterfactual removals
exactly, the inhibition is directly interpretable, and it doubles as a
control surface. Suppressing a
targeted sense cuts gender bias by up to $39.7\%$ at negligible cost. The
senses are also better representations, improving sentence similarity by
up to $33\%$ over the softmax Backpack.

A ladder of combination rules then sharpens what the architecture costs:
making the readout nonlinear gains $0.43$ ppl, improves no probe beyond
noise and
invalidates two thirds of the decomposition, while the $82\%$ of the tax
that survives is the depth displaced by the sense dictionary's parameters;
at matched depth the differential Backpack trails a Transformer by $0.14$
ppl. Since exact attribution requires only a linear readout over a fixed
inspectable basis, and nothing of how the coefficients are computed, the
natural next step is a global overcomplete dictionary. Non-negativity was never the price of interpretability, and
neither is linearity.

% ACL: content is limited to 8 pages; Limitations and everything after it
% are unlimited and begin on a new page.
\clearpage
\section*{Limitations}
\paragraph{Scale and replication.}
Our experiments span $70$M to $350$M parameters. The advantage persists
across that range but shrinks with scale, from $4.4\%$ relative at $70$M
to $2.9\%$ at $170$M and $2.5\%$ at $350$M, and the tax reduction from
$31\%$ to $21\%$; whether it survives at billion-parameter scale is open.
The WikiText-103 and $170$M OpenWebText comparisons are replicated over
three seeds per arm; the $350$M runs, the Transformer baselines, the
sense-count sweep and the control experiment are single-seed. The depth-matched Transformer has $27\%$ fewer parameters than the
Backpack, so the two Transformer baselines bound the dictionary's cost
from either side; a joint sweep over depth and dictionary size would give
the full trade-off.

\paragraph{Size and cost of the effect.}
The perplexity gains, while consistent, are modest in absolute terms:
differential contextualization reduces the Backpack's gap to a Transformer
but does not close it, and the contribution is best understood as lowering
the cost of an interpretable architecture rather than advancing the state
of the art in perplexity. Differential contextualization adds a second attention map at the sense
layer, which costs under $1\%$ in parameters but $12$--$16\%$ of training
throughput (Appendix~\ref{app:hparams}).

\paragraph{Interpretability evidence.}
The sense-level interpretations, though exact by construction, are cleaner
for some words than others; our disambiguation analysis shows a clear case
and does not claim every sense is human-labelable, and the inhibition
ablation (\S\ref{sec:inhibit}) shows that inhibition, while essential,
is not polysemy-specific, so the disambiguation reading describes
instances rather than its dataset-level role. That evidence is
qualitative: a zero-shot WiC probe (\S\ref{sec:wic}) shows no measurable
advantage for signed sense weights, though at this model scale even a
contextual hidden-state baseline barely exceeds the majority class, so the
experiment does not decide the question either way. The control result (\S\ref{sec:control}) is single-seed at each scale,
and the reversal of the frontier at $70$M is unexplained.

\paragraph{Scope.}
All experiments are on English, and signed weights forgo the ``conic
selection'' reading of the original Backpack even though exact per-sense
attribution is retained.

\section*{Ethics Statement}
The debiasing experiment (\S\ref{sec:control}) measures gender bias as
the gap between the log-probabilities of \textit{he} and \textit{she}
after $20$ profession nouns in six templates. This binary, English,
pronoun-based measure is a standard probe of one salient bias and not a
measure of fairness: it omits non-binary identities, other languages, and
the many forms bias takes beyond pronoun prediction, and reducing it does
not make a model unbiased. Sense suppression is an inference-time edit
that can remove a targeted association; the same mechanism could be used
to insert one, which is a reason to prefer architectures whose edits are
inspectable, as the linear readout makes ours. All models are trained on
public corpora (WikiText-103 and OpenWebText) that contain the biases of
their sources; we release no new data. Training used a single H200 GPU
for roughly one GPU-week in total.

\section*{AI Assistance}
An AI coding and writing assistant (Claude, Anthropic) was used under the
authors' direction to implement the experimental code from the authors'
specification and to produce analysis scripts and figures. The research
idea, the experimental design, the execution of all experiments, the
interpretation of the results, and the claims made are the authors' own,
and all numbers were verified against the experimental records.

\bibliography{references}

@inproceedings{hewitt-etal-2023-backpack,
  title     = {Backpack Language Models},
  author    = {Hewitt, John and Thickstun, John and Manning, Christopher D. and Liang, Percy},
  booktitle = {Proceedings of the 61st Annual Meeting of the Association for Computational Linguistics (Volume 1: Long Papers)},
  year      = {2023},
  address   = {Toronto, Canada},
  publisher = {Association for Computational Linguistics},
  pages     = {9103--9125},
  url       = {https://aclanthology.org/2023.acl-long.506},
}

@inproceedings{ye2024differential,
  title     = {Differential Transformer},
  author    = {Ye, Tianzhu and Dong, Li and Xia, Yuqing and Sun, Yutao and Zhu, Yi and Huang, Gao and Wei, Furu},
  booktitle = {The Thirteenth International Conference on Learning Representations (ICLR)},
  year      = {2025},
  url       = {https://arxiv.org/abs/2410.05258},
}

@inproceedings{pilehvar-camacho-collados-2019-wic,
  title     = {{W}i{C}: the Word-in-Context Dataset for Evaluating Context-Sensitive Meaning Representations},
  author    = {Pilehvar, Mohammad Taher and Camacho-Collados, Jose},
  booktitle = {Proceedings of the 2019 Conference of the North American Chapter of the Association for Computational Linguistics: Human Language Technologies, Volume 1 (Long and Short Papers)},
  year      = {2019},
  address   = {Minneapolis, Minnesota},
  publisher = {Association for Computational Linguistics},
  pages     = {1267--1273},
  url       = {https://aclanthology.org/N19-1128},
}

@inproceedings{sun2023chinese,
  title     = {Character-Level {C}hinese Backpack Language Models},
  author    = {Sun, Hao and Hewitt, John},
  booktitle = {Proceedings of the 6th BlackboxNLP Workshop: Analyzing and Interpreting Neural Networks for NLP},
  year      = {2023},
  address   = {Singapore},
  publisher = {Association for Computational Linguistics},
  pages     = {106--119},
  url       = {https://aclanthology.org/2023.blackboxnlp-1.8},
}

@article{hewitt2024model,
  title   = {Model Editing with Canonical Examples},
  author  = {Hewitt, John and Chen, Sarah and Xie, Lanruo Lora and Adams, Edward and Liang, Percy and Manning, Christopher D.},
  journal = {arXiv preprint arXiv:2402.06155},
  year    = {2024},
  url     = {https://arxiv.org/abs/2402.06155},
}

@article{cruz2026sense,
  title   = {Sense Representations Are Inducible Interfaces},
  author  = {Cruz, Jan Christian Blaise and Aji, Alham Fikri},
  journal = {arXiv preprint arXiv:2605.28669},
  year    = {2026},
  url     = {https://arxiv.org/abs/2605.28669},
}

@article{lv2024cog,
  title   = {More Expressive Attention with Negative Weights},
  author  = {Lv, Ang and Xie, Ruobing and Li, Shuaipeng and Liao, Jiayi and Sun, Xingwu and Kang, Zhanhui and Wang, Di and Yan, Rui},
  journal = {arXiv preprint arXiv:2411.07176},
  year    = {2024},
  url     = {https://arxiv.org/abs/2411.07176},
}

@inproceedings{kobyzev-etal-2025-integral,
  title     = {Integral Transformer: Denoising Attention, Not Too Much Not Too Little},
  author    = {Kobyzev, Ivan and Ghaddar, Abbas and Hu, Dingtao and Chen, Boxing},
  booktitle = {Proceedings of the 2025 Conference on Empirical Methods in Natural Language Processing},
  year      = {2025},
  pages     = {2337--2354},
  url       = {https://aclanthology.org/2025.emnlp-main.118},
}

@article{huang2026threshold,
  title   = {Threshold Differential Attention for Sink-Free, Ultra-Sparse, and Non-Dispersive Language Modeling},
  author  = {Huang, Xingyue and Ding, Xueying and Ju, Mingxuan and Liu, Yozen and Shah, Neil and Zhao, Tong},
  journal = {arXiv preprint arXiv:2601.12145},
  year    = {2026},
  note    = {To appear at ACL 2026},
  url     = {https://arxiv.org/abs/2601.12145},
}

@inproceedings{afzali2026backpack,
  title     = {Controlling Gender Bias in Retrieval via a Backpack Architecture},
  author    = {Afzali, Amirabbas and Velae, Amirreza and Ahmadi, Iman and Aliannejadi, Mohammad},
  booktitle = {Proceedings of the 48th European Conference on Information Retrieval (ECIR)},
  year      = {2026},
  note      = {arXiv:2511.00875},
  url       = {https://arxiv.org/abs/2511.00875},
}

@article{cruz2026multilingual,
  title   = {Multilinguality as Sense Adaptation},
  author  = {Cruz, Jan Christian Blaise and Adelani, David Ifeoluwa and Aji, Alham Fikri},
  journal = {arXiv preprint arXiv:2601.10310},
  year    = {2026},
  url     = {https://arxiv.org/abs/2601.10310},
}

@article{diff-gated,
  title   = {Differential Gated Self-Attention},
  author  = {Lygizou, Elpiniki Maria and Farsang, M{\'o}nika and Grosu, Radu},
  journal = {arXiv preprint arXiv:2505.24054},
  year    = {2025},
  url     = {https://arxiv.org/abs/2505.24054},
}

@article{shared-diff,
  title   = {Shared {DIFF} Transformer},
  author  = {Cang, Yueyang and Liu, Yuhang and Zhang, Xiaoteng and Shi, Li and Que, Wenge},
  journal = {arXiv preprint arXiv:2501.17900},
  year    = {2025},
  url     = {https://arxiv.org/abs/2501.17900},
}

@inproceedings{understanding-diff,
  title     = {Understanding Differential Transformer Unchains Pretrained Self-Attentions},
  author    = {Kong, Chaerin and Jang, Jiho and Kwak, Nojun},
  booktitle = {Advances in Neural Information Processing Systems (NeurIPS)},
  year      = {2025},
  note      = {arXiv:2505.16333},
  url       = {https://arxiv.org/abs/2505.16333},
}

@article{elhage2022superposition,
  title   = {Toy Models of Superposition},
  author  = {Elhage, Nelson and Hume, Tristan and Olsson, Catherine and Schiefer, Nicholas and Henighan, Tom and Kravec, Shauna and Hatfield-Dodds, Zac and Lasenby, Robert and Drain, Dawn and Chen, Carol and Grosse, Roger and McCandlish, Sam and Kaplan, Jared and Amodei, Dario and Wattenberg, Martin and Olah, Christopher},
  journal = {Transformer Circuits Thread},
  year    = {2022},
  url     = {https://transformer-circuits.pub/2022/toy_model/index.html},
}

@article{elhage2021framework,
  title   = {A Mathematical Framework for Transformer Circuits},
  author  = {Elhage, Nelson and Nanda, Neel and Olsson, Catherine and Henighan, Tom and Joseph, Nicholas and Mann, Ben and Askell, Amanda and Bai, Yuntao and Chen, Anna and Conerly, Tom and DasSarma, Nova and Drain, Dawn and Ganguli, Deep and Hatfield-Dodds, Zac and Hernandez, Danny and Jones, Andy and Kernion, Jackson and Lovitt, Liane and Ndousse, Kamal and Amodei, Dario and Brown, Tom and Clark, Jack and Kaplan, Jared and McCandlish, Sam and Olah, Chris},
  journal = {Transformer Circuits Thread},
  year    = {2021},
  url     = {https://transformer-circuits.pub/2021/framework/index.html},
}

@inproceedings{huben2024sparse,
  title     = {Sparse Autoencoders Find Highly Interpretable Features in Language Models},
  author    = {Huben, Robert and Cunningham, Hoagy and Smith, Logan and Ewart, Aidan and Sharkey, Lee},
  booktitle = {The Twelfth International Conference on Learning Representations (ICLR)},
  year      = {2024},
  url       = {https://arxiv.org/abs/2309.08600},
}

@inproceedings{reisinger-mooney-2010-multi,
  title     = {Multi-Prototype Vector-Space Models of Word Meaning},
  author    = {Reisinger, Joseph and Mooney, Raymond J.},
  booktitle = {Human Language Technologies: The 2010 Annual Conference of the North American Chapter of the Association for Computational Linguistics},
  year      = {2010},
  pages     = {109--117},
  url       = {https://aclanthology.org/N10-1013},
}

@inproceedings{neelakantan-etal-2014-efficient,
  title     = {Efficient Non-parametric Estimation of Multiple Embeddings per Word in Vector Space},
  author    = {Neelakantan, Arvind and Shankar, Jeevan and Passos, Alexandre and McCallum, Andrew},
  booktitle = {Proceedings of the 2014 Conference on Empirical Methods in Natural Language Processing (EMNLP)},
  year      = {2014},
  pages     = {1059--1069},
  url       = {https://aclanthology.org/D14-1113},
}

@article{mosolova2026wsi,
  title   = {In the {LLM} era, Word Sense Induction remains unsolved},
  author  = {Mosolova, Anna and Candito, Marie and Ramisch, Carlos},
  journal = {arXiv preprint arXiv:2603.11686},
  year    = {2026},
  url     = {https://arxiv.org/abs/2603.11686},
}

@article{camacho-collados-pilehvar-2018-survey,
  title   = {From Word to Sense Embeddings: A Survey on Vector Representations of Meaning},
  author  = {Camacho-Collados, Jose and Pilehvar, Mohammad Taher},
  journal = {Journal of Artificial Intelligence Research},
  volume  = {63},
  pages   = {743--788},
  year    = {2018},
}

@inproceedings{jain-wallace-2019-attention,
  title     = {Attention is not Explanation},
  author    = {Jain, Sarthak and Wallace, Byron C.},
  booktitle = {Proceedings of the 2019 Conference of the North American Chapter of the Association for Computational Linguistics: Human Language Technologies, Volume 1 (Long and Short Papers)},
  year      = {2019},
  pages     = {3543--3556},
  url       = {https://aclanthology.org/N19-1357},
}

@inproceedings{wiegreffe-pinter-2019-attention,
  title     = {Attention is not not Explanation},
  author    = {Wiegreffe, Sarah and Pinter, Yuval},
  booktitle = {Proceedings of the 2019 Conference on Empirical Methods in Natural Language Processing and the 9th International Joint Conference on Natural Language Processing (EMNLP-IJCNLP)},
  year      = {2019},
  pages     = {11--20},
  url       = {https://aclanthology.org/D19-1002},
}

@inproceedings{merity2017pointer,
  title     = {Pointer Sentinel Mixture Models},
  author    = {Merity, Stephen and Xiong, Caiming and Bradbury, James and Socher, Richard},
  booktitle = {International Conference on Learning Representations (ICLR)},
  year      = {2017},
  url       = {https://arxiv.org/abs/1609.07843},
}

@article{radford2019language,
  title   = {Language Models are Unsupervised Multitask Learners},
  author  = {Radford, Alec and Wu, Jeffrey and Child, Rewon and Luan, David and Amodei, Dario and Sutskever, Ilya},
  journal = {OpenAI blog},
  volume  = {1},
  number  = {8},
  year    = {2019},
}

@inproceedings{vaswani2017attention,
  title     = {Attention Is All You Need},
  author    = {Vaswani, Ashish and Shazeer, Noam and Parmar, Niki and Uszkoreit, Jakob and Jones, Llion and Gomez, Aidan N. and Kaiser, Lukasz and Polosukhin, Illia},
  booktitle = {Advances in Neural Information Processing Systems 30},
  year      = {2017},
}

@article{hill2015simlex,
  title   = {{S}im{L}ex-999: Evaluating Semantic Models With (Genuine) Similarity Estimation},
  author  = {Hill, Felix and Reichart, Roi and Korhonen, Anna},
  journal = {Computational Linguistics},
  volume  = {41},
  number  = {4},
  pages   = {665--695},
  year    = {2015},
  url     = {https://aclanthology.org/J15-4004},
}

@inproceedings{gerz-etal-2016-simverb,
  title     = {{S}im{V}erb-3500: A Large-Scale Evaluation Set of Verb Similarity},
  author    = {Gerz, Daniela and Vuli{\'c}, Ivan and Hill, Felix and Reichart, Roi and Korhonen, Anna},
  booktitle = {Proceedings of the 2016 Conference on Empirical Methods in Natural Language Processing},
  year      = {2016},
  pages     = {2173--2182},
  url       = {https://aclanthology.org/D16-1235},
}

@article{rubenstein1965contextual,
  title   = {Contextual Correlates of Synonymy},
  author  = {Rubenstein, Herbert and Goodenough, John B.},
  journal = {Communications of the ACM},
  volume  = {8},
  number  = {10},
  pages   = {627--633},
  year    = {1965},
}

@inproceedings{finkelstein2001placing,
  title     = {Placing Search in Context: The Concept Revisited},
  author    = {Finkelstein, Lev and Gabrilovich, Evgeniy and Matias, Yossi and Rivlin, Ehud and Solan, Zach and Wolfman, Gadi and Ruppin, Eytan},
  booktitle = {Proceedings of the 10th International Conference on World Wide Web},
  pages     = {406--414},
  year      = {2001},
}

@inproceedings{socher-etal-2013-recursive,
  title     = {Recursive Deep Models for Semantic Compositionality Over a Sentiment Treebank},
  author    = {Socher, Richard and Perelygin, Alex and Wu, Jean and Chuang, Jason and Manning, Christopher D. and Ng, Andrew and Potts, Christopher},
  booktitle = {Proceedings of the 2013 Conference on Empirical Methods in Natural Language Processing},
  year      = {2013},
  pages     = {1631--1642},
  url       = {https://aclanthology.org/D13-1170},
}

@inproceedings{dolan-brockett-2005-automatically,
  title     = {Automatically Constructing a Corpus of Sentential Paraphrases},
  author    = {Dolan, William B. and Brockett, Chris},
  booktitle = {Proceedings of the Third International Workshop on Paraphrasing (IWP2005)},
  year      = {2005},
  url       = {https://aclanthology.org/I05-5002},
}

@inproceedings{cer-etal-2017-semeval,
  title     = {{S}em{E}val-2017 Task 1: Semantic Textual Similarity Multilingual and Crosslingual Focused Evaluation},
  author    = {Cer, Daniel and Diab, Mona and Agirre, Eneko and Lopez-Gazpio, I{\~n}igo and Specia, Lucia},
  booktitle = {Proceedings of the 11th International Workshop on Semantic Evaluation (SemEval-2017)},
  year      = {2017},
  pages     = {1--14},
  url       = {https://aclanthology.org/S17-2001},
}

@inproceedings{silveira-etal-2014-gold,
  title     = {A Gold Standard Dependency Corpus for {E}nglish},
  author    = {Silveira, Natalia and Dozat, Timothy and de Marneffe, Marie-Catherine and Bowman, Samuel R. and Connor, Miriam and Bauer, John and Manning, Christopher D.},
  booktitle = {Proceedings of the Ninth International Conference on Language Resources and Evaluation (LREC'14)},
  year      = {2014},
  pages     = {2897--2904},
  url       = {https://aclanthology.org/L14-1067},
}

@inproceedings{devlin-etal-2019-bert,
  title     = {{BERT}: Pre-training of Deep Bidirectional Transformers for Language Understanding},
  author    = {Devlin, Jacob and Chang, Ming-Wei and Lee, Kenton and Toutanova, Kristina},
  booktitle = {Proceedings of the 2019 Conference of the North American Chapter of the Association for Computational Linguistics: Human Language Technologies, Volume 1 (Long and Short Papers)},
  year      = {2019},
  pages     = {4171--4186},
  url       = {https://aclanthology.org/N19-1423},
}

@inproceedings{gao-etal-2021-simcse,
  title     = {{S}im{CSE}: Simple Contrastive Learning of Sentence Embeddings},
  author    = {Gao, Tianyu and Yao, Xingcheng and Chen, Danqi},
  booktitle = {Proceedings of the 2021 Conference on Empirical Methods in Natural Language Processing},
  year      = {2021},
  pages     = {6894--6910},
  url       = {https://aclanthology.org/2021.emnlp-main.552},
}

@misc{gokaslan2019openwebtext,
  title  = {{O}pen{W}eb{T}ext Corpus},
  author = {Gokaslan, Aaron and Cohen, Vanya},
  year   = {2019},
  url    = {http://Skylion007.github.io/OpenWebTextCorpus},
}

\appendix

\section{Proofs}
\label{app:proofs}
Both propositions are stated in \S\ref{sec:theory}; we give the arguments
here in full.

\begin{proof}[Proof of Proposition~\ref{prop:reach}]
Softmax rows are convex weights, giving the first claim. For the second,
necessity: summing $b = p - \lambda_\ell q$ gives $\sum_j b_j =
1-\lambda_\ell$. Since $\lambda_\ell q_j \ge 0$, every coordinate satisfies
$p_j = b_j + \lambda_\ell q_j \ge b_j$, hence $\max(b_j,0) \le p_j$
(trivially when $b_j \le 0$) and $\sum_j \max(b_j,0) \le \sum_j p_j = 1$.
Writing $\sum_j \max(-b_j,0) = \sum_j \max(b_j,0) - \sum_j b_j$ and using
the sum constraint gives $\sum_j \max(-b_j,0) \le 1 - (1-\lambda_\ell) =
\lambda_\ell$. Sufficiency: set
$q_j = \max(-b_j,0)/\lambda_\ell$, place the remaining mass
$1 - \sum_j q_j \ge 0$ on any coordinate, and take $p = b + \lambda_\ell q$.
Then $p_j \ge 0$ by construction and $\sum_j p_j = (1-\lambda_\ell) +
\lambda_\ell = 1$. For \eqref{eq:l1ball}, dividing \eqref{eq:reach} by
$1-\lambda_\ell > 0$ gives $\sum_j a_j = 1$ and $\sum_j \max(-a_j,0) \le
\lambda_\ell/(1-\lambda_\ell)$; on the hyperplane $\sum_j a_j = 1$,
$\|a\|_1 = \sum_j a_j + 2\sum_j\max(-a_j,0)$, so
$\|a\|_1 \le 1 + 2\lambda_\ell/(1-\lambda_\ell) =
(1+\lambda_\ell)/(1-\lambda_\ell)$. The bound is attained by $p = e_1$,
$q = e_2$, for which $a = (1, -\lambda_\ell)/(1-\lambda_\ell)$. At
$\lambda_\ell = 0$ this collapses to
$\|a\|_1 = 1$ with $\sum_j a_j = 1$, i.e.\ the simplex. The equivalence
holds for a single position as well; $i \ge 2$ is needed only for the
strict inclusion in \eqref{eq:l1ball}, which requires a coordinate that can
go negative. The familiar box bounds $b_j \in [-\lambda_\ell, 1]$ are
implied by \eqref{eq:reach} but strictly weaker: $b = (1, 0.8, -0.8, -0.8)$
with $\lambda_\ell = 0.8$ lies in the box and has the correct sum, yet its
positive part totals $1.8 > 1$, so it is unreachable (verified against an
exact linear program in the released code).
\end{proof}

\begin{proof}[Proof of Proposition~\ref{prop:sep}]
The single-token constraint fixes $C(x_i) = u$ for softmax, whose lone
weight is $1$, and $C(x_i) = u/(1-\lambda)$ for the differential Backpack,
whose lone weight is $1-\lambda$; each model realizes the same pinned
output with its own sense scaling. Softmax then gives
$o = \alpha u + (1-\alpha)v$, $\alpha \in [0,1]$; in the $(u,v)$ basis the
target is $(1+s,-s)$ and the reachable segment is $\{(\alpha, 1-\alpha)\}$,
so the squared distance is $\min_{\alpha \in [0,1]} \big[(1+s-\alpha)^2 +
(1-\alpha+s)^2\big] = \min_{\alpha \in [0,1]} 2(1+s-\alpha)^2 = 2s^2$,
attained at $\alpha = 1$. For the differential Backpack the output is
$o = (b_1 u + b_2 v)/(1-\lambda)$, so the target requires
$b = (1-\lambda)(1+s, -s)$, which has the correct sum $1-\lambda$ and
negative part $(1-\lambda)s$. By \eqref{eq:reach} it is reachable exactly
when $(1-\lambda)s \le \lambda$; since $1-\lambda > 0$ this is
$s \le \lambda/(1-\lambda)$, and rearranging $s - \lambda s \le \lambda$
gives $\lambda \ge s/(1+s)$. (With a single negative coordinate the
constraint \eqref{eq:reach} reduces to the coordinate bound
$b_j \ge -\lambda$.)
\end{proof}

\section{Hyperparameters and Reproducibility}
\label{app:hparams}
All models use pre-LN Transformer blocks, tied input/output embeddings,
learned absolute positions, and dropout $0.1$. WikiText-103 runs: $d{=}512$,
$8$ layers, $8$ heads, $k{=}16$ senses, context $512$, per-sense q/k
dimension $32$; batch $32$ with gradient accumulation $4$; AdamW
($\beta{=}0.9,0.95$, weight decay $0.1$), peak LR $3\times10^{-4}$ with
$200$ warmup steps and cosine decay to $3\times10^{-5}$; $20$k steps in
bf16. Transformer baselines use $14$ layers to match parameters and $8$
to match depth ($51$M). The
OpenWebText runs use gradient accumulation $16$ and $50$k steps, with
$d{=}768$, $12$ layers and $12$ heads at $170$M, and $d{=}1024$, $16$ heads
and $18$ layers ($24$ for the parameter-matched Transformer) at $350$M. The
differential $\lambda$ uses the reparameterization of
\citet{ye2024differential} with $\lambda_{\mathrm{init}}{=}0.5$ at the sense
layer.

\paragraph{Cost of the second map.}
Differential contextualization adds one query/key projection at the sense
layer and nothing else: $0.53$M, $0.79$M and $1.05$M parameters at $70$M,
$170$M and $350$M ($0.76\%$, $0.48\%$ and $0.30\%$ of the model). By
operation count the second map, a second attention pass over the sense
vectors, is $5.7\%$, $3.6\%$ and $2.3\%$ of the forward FLOPs at the three
scales. Measured training throughput falls by more, $15.6\%$, $14.4\%$
and $11.7\%$ relative to the softmax Backpack on the same fused-attention
path, because attention over $k{=}16$ sense heads with $d$-dimensional
values is memory-bound rather than compute-bound. Relative to the
parameter-matched Transformer, the softmax Backpack trains $7\%$ slower at
$70$M and $4\%$ at $350$M, the differential Backpack $22\%$ and $15\%$.
Inference carries the same two passes; inspecting the weights is
unchanged, since the two maps are read off the same forward pass. Peak
memory was not measured. Throughput of the control rungs (one-hot, sum,
gain, $\lambda{=}0$, Cog, unconstrained) is not comparable, as they use
the explicit, unfused weight computation.

\begin{figure}[t]
  \centering
  \includegraphics[width=0.82\columnwidth]{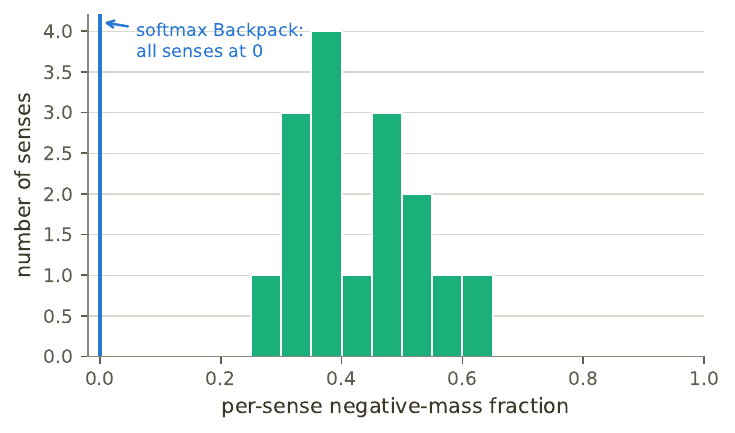}
  \caption{Per-sense negative-mass fraction of the differential Backpack
  in the main configuration ($k{=}16$); $0$ is purely excitatory, $1$ purely
  inhibitory.}
  \label{fig:neghist}
\end{figure} 
\section{Faithfulness and Inhibition Ablation}
\label{app:inhibit}
\begin{table}[t]
\centering
\small
\setlength{\tabcolsep}{3pt}
\begin{tabular}{lcccc}
\toprule
Rule & $\rho$ & rel.\ err. & compr. & suffic. \\
\midrule
one-hot               & 1.000 & $10^{-6}$ & 5.97 & $+0.27$ \\
softmax               & 1.000 & $10^{-6}$ & 3.49 & $-1.43$ \\
diff Backpack (ours)  & 1.000 & $10^{-6}$ & 3.74 & $-1.22$ \\
unconstr.\ weights & 1.000 & $10^{-6}$ & 1.14 & $-4.88$ \\
Cog Attention      & 1.000 & $10^{-6}$ & 3.63 & $-1.39$ \\
\midrule
softmax\,+\,MLP       & 0.77 & 0.48 & 3.39 & $-1.45$ \\
diff\,+\,MLP          & 0.82 & 0.44 & 3.10 & $-1.26$ \\
\midrule
ten random pairs      & --- & --- & 0.01 & $-7.5$ \\
\bottomrule
\end{tabular}
\caption{Counterfactual faithfulness of the per-sense attribution on
WikiText-103 ($300$ positions): Spearman correlation and median relative
error between attributed and actual logit change under single removals of
the ten top pairs, and the change in target log-probability (nats) when
those pairs are removed jointly (comprehensiveness) or kept alone
(sufficiency).}
\label{tab:faithful}
\end{table}

Table~\ref{tab:faithful} gives the counterfactual faithfulness numbers of
\S\ref{sec:faithful}.
\begin{figure*}[t]
  \centering
  \includegraphics[width=\textwidth]{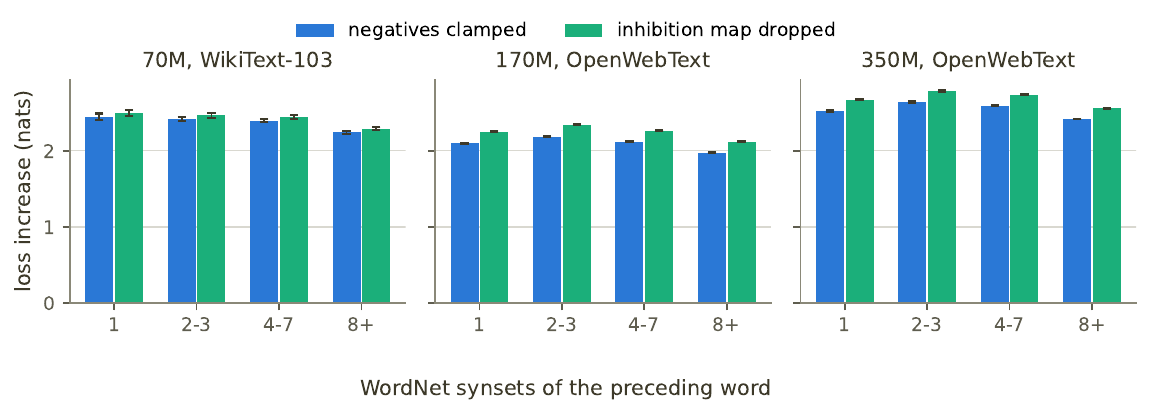}
  \caption{Per-token loss increase when inhibition is removed at inference
  from a trained differential Backpack, by WordNet polysemy of the
  preceding whole word, for the two edits of \S\ref{sec:inhibit}; bars
  are means with $95\%$ intervals over all validation tokens.}
  \label{fig:inhibit}
\end{figure*}
Both edits act on the explicit weights $A^\ell = A_1 - \lambda_\ell A_2$
and keep every row's mass $1-\lambda_\ell$, so only the sign structure is
removed. Polysemy is the number of WordNet synsets of the preceding token
when it is a complete alphabetic word ($53\%$ of tokens are word pieces,
punctuation or words absent from WordNet and are reported separately; the
loss increase there is larger still, $3.0$--$3.3$ nats). Applied to the
softmax Backpack both edits are identities and the measured change is
exactly zero. The loss increase is $2.2$--$2.8$ nats after every class of
lexical word at all three scales and decreases mildly with polysemy
(Figure~\ref{fig:inhibit}); relative to the base loss it is flat, about
$0.73$ throughout.

\section{Scale: OpenWebText at 170M and 350M}
\label{sec:owt}
Table~\ref{tab:scale} collects the equal-budget comparisons across scale;
every OpenWebText model is trained on the same $13.1$B tokens. The
advantage persists to $350$M and the three seeds at $170$M put it at
$9$ standard errors, but the relative gain shrinks with scale and so does
the tax reduction, from $31\%$ at $70$M to $21\%$ at $350$M, while
inhibition usage does not (negative mass $0.42$--$0.45$, mean $\lambda$
$0.84$--$0.88$). Table~\ref{tab:probes} gives every probe at every scale.
The profile is stable: the differential Backpack's senses lead the softmax Backpack's on STS-B, SST-2 and sense clustering and trail on POS at $70$M, $170$M and $350$M, with
the STS-B margin shrinking from $0.12$ to $0.06$; the $354$M
Transformer's hidden state reaches $0.572$ on STS-B. For reference, the
STS-B protocol gives $0.428$ for GPT-2, $0.593$ for BERT-base
\citep{devlin-etal-2019-bert} and $0.856$ for SimCSE
\citep{gao-etal-2021-simcse}, so the sense outputs sit between GPT-2 and
BERT-base. MRPC probes stay at or below the majority class ($0.684$) for
every model below $350$M and reach it only for the $350$M Transformer
($0.711$) and the differential Backpack's senses ($0.684$), so the task does not separate
models at these scales. The depth-matched $8$-layer Transformer matches
the $14$-layer one on every probe (STS-B $0.484$ vs.\ $0.488$, SST-2
$0.744$ vs.\ $0.753$) while trailing it by $2.29$ ppl, another sign that
the probes measure something perplexity does not.

\begin{table}[t]
\centering
\small
\setlength{\tabcolsep}{4pt}
\begin{tabular}{lcccc}
\toprule
 & neg. & entropy & sparsity & SL999/WS353 \\
 & mass & of $|A|$ & @1e-3 & (best $\rho$) \\
\midrule
softmax & 0.00 & 0.70 & 0.66 & 0.42 / 0.70 \\
diff    & 0.45 & 0.78 & 0.49 & 0.44 / 0.73 \\
\bottomrule
\end{tabular}
\caption{Sense-weight structure and lexical quality on
WikiText-103 (main configuration): negative weight mass, normalized entropy and sparsity of
$|A|$, and best-sense Spearman $\rho$ on SimLex-999 / WordSim-353
(raw-embedding baselines $0.37$ / $0.42$--$0.45$).}
\label{tab:sense}
\end{table}

\begin{figure*}[t]
  \centering
  \includegraphics[width=\textwidth]{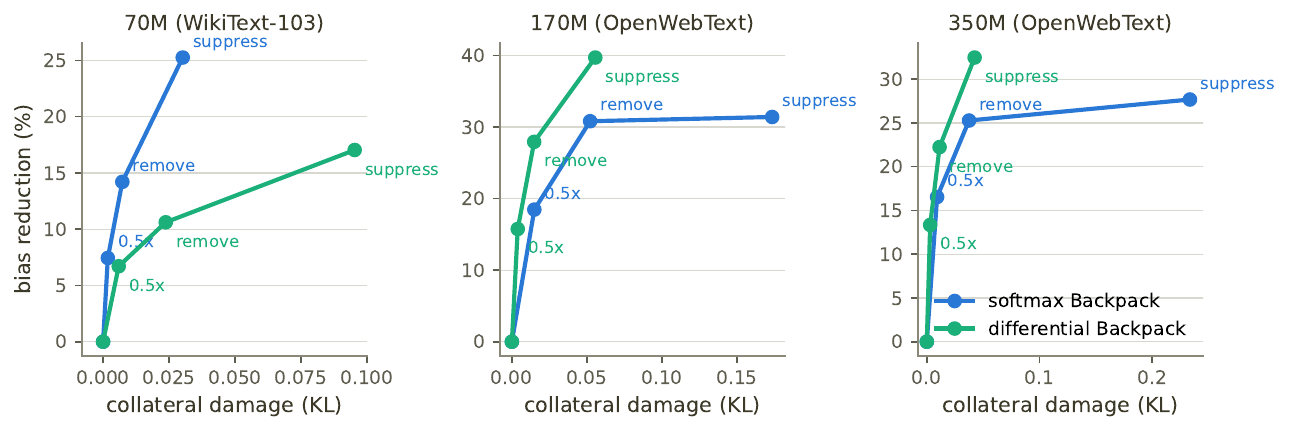}
  \caption{Bias against collateral damage (KL) for the
  three sense interventions at $70$M, $170$M and $350$M (left to right);
  upper-left is better.}
  \label{fig:control}
\end{figure*}

\begin{table}[t]
\centering
\small
\begin{tabular}{llcccc}
\toprule
 & & none & $0.5\times$ & remove & suppress \\
\midrule
\multirow{4}{*}{\rotatebox{90}{70M}}
 & softmax bias & 1.14 & 1.06 & 0.98 & 0.86 \\
 & \quad KL     & ---  & .002 & .007 & .030 \\
 & diff bias    & 1.77 & 1.65 & 1.58 & 1.47 \\
 & \quad KL     & ---  & .006 & .024 & .095 \\
\midrule
\multirow{4}{*}{\rotatebox{90}{170M}}
 & softmax bias & 1.74 & 1.42 & 1.20 & 1.19 \\
 & \quad KL     & ---  & .015 & .052 & .173 \\
 & diff bias    & 1.92 & 1.62 & 1.38 & \textbf{1.16} \\
 & \quad KL     & ---  & .004 & .015 & \textbf{.056} \\
\midrule
\multirow{4}{*}{\rotatebox{90}{350M}}
 & softmax bias & 1.29 & 1.08 & 0.96 & 0.93 \\
 & \quad KL     & ---  & .009 & .037 & .233 \\
 & diff bias    & 1.49 & 1.29 & 1.16 & \textbf{1.01} \\
 & \quad KL     & ---  & .003 & .011 & \textbf{.042} \\
\bottomrule
\end{tabular}
\caption{Gender bias ($\downarrow$) and collateral damage
KL ($\downarrow$) under each sense intervention, over $20$ profession
nouns, at $70$M, $170$M and $350$M.}
\label{tab:control}
\end{table}

\begin{table}[t]
\centering
\small
\setlength{\tabcolsep}{3pt}
\footnotesize
\begin{tabular}{lccccc}
\toprule
Scale & Transf. & softmax & diff & gap & cut \\
\midrule
70M WT103  & 21.09 & 24.60 & 23.52 & 1.08 (4.4\%) & 31\% \\
170M OWT   & ---   & 23.84 & 23.15 & 0.70 (2.9\%) & ---  \\
350M OWT   & 18.02 & 20.48 & 19.96 & 0.52 (2.5\%) & 21\% \\
\bottomrule
\end{tabular}
\caption{Exact validation perplexity across scale at equal parameters and
token budget; gap is softmax minus diff and cut the reduction of the tax
(the gap to the parameter-matched Transformer). The 70M and 170M Backpack
entries are three-seed means (s.d.\ $0.12$/$0.04$ and $0.13$/$0.03$ for
softmax/diff); the rest are single runs.}
\label{tab:scale}
\end{table}

\begin{table}[t]
\centering
\small
\setlength{\tabcolsep}{2pt}
\footnotesize
\begin{tabular}{lccccc}
\toprule
Repr. & POS & STS & SST-2 & MRPC & ARI \\
\midrule
Transf.\ ($h$), 70M       & 0.859 & 0.488 & 0.753 & 0.662 & 0.174 \\
Transf.\ 8L ($h$), 70M    & 0.860 & 0.484 & 0.744 & 0.674 & 0.160 \\
DIFF Transf.\ ($h$), 70M  & 0.859 & 0.499 & 0.743 & 0.654 & 0.203 \\
softmax ($o$), 70M               & 0.844 & 0.372 & 0.718 & 0.625 & 0.151 \\
diff ($o$), 70M                  & 0.815 & 0.495 & 0.731 & 0.613 & 0.174 \\
Cog ($o$), 70M            & 0.836 & 0.422 & 0.736 & 0.632 & 0.163 \\
diff+MLP ($o$), 70M       & 0.838 & 0.485 & 0.749 & 0.662 & 0.168 \\
\midrule
softmax ($o$), 170M              & 0.850 & 0.466 & 0.773 & 0.657 & 0.193 \\
diff ($o$), 170M                 & 0.834 & 0.554 & 0.792 & 0.679 & 0.215 \\
\midrule
Transf.\ ($h$), 350M      & 0.869 & 0.572 & 0.822 & 0.711 & 0.256 \\
softmax ($o$), 350M              & 0.855 & 0.476 & 0.786 & 0.676 & 0.194 \\
diff ($o$), 350M                 & 0.844 & 0.539 & 0.792 & 0.684 & 0.204 \\
\bottomrule
\end{tabular}
\caption{Frozen-representation probes at every scale: POS accuracy, STS-B
Spearman, SST-2 and MRPC probe accuracy (majority class $0.509$ and
$0.684$), and sense-clustering ARI.}
\label{tab:probes}
\end{table}

\section{Sense Specialization}
\label{sec:spec}
Running our battery on the released Backpack-GPT2 gives an attribution
residual of $4\!\times\!10^{-7}$ and negative weight mass exactly zero. It
reaches $19.6$ ppl on OpenWebText with roughly $4\times$ our token budget;
our $350$M differential Backpack reaches $19.96$ at a quarter of the tokens
and twice the parameters, and our $170$M model reaches comparable
sense-output quality on STS-B ($0.554$ vs.\ $0.574$). The battery also
reproduces the profile of Table~4 in \citet{hewitt-etal-2023-backpack}: min-sense similarity is
strongest on SimLex-999 and a single sense on RG-65 and WordSim-353, as
reported, while our absolute values run $0.02$--$0.07$ higher (e.g.\
$0.568$ vs.\ $0.540$ min-sense on SimLex-999), which we attribute to
vocabulary coverage and to selecting the best sense per dataset.

The richer weight usage is matched by richer senses
(Table~\ref{tab:spec}). We score sense vectors against human similarity
judgments on SimLex-999 \citep{hill2015simlex}, SimVerb-3500
\citep{gerz-etal-2016-simverb}, RG-65 \citep{rubenstein1965contextual} and
WordSim-353 \citep{finkelstein2001placing}. Sense vectors far exceed the
raw-embedding baseline on both lexical-similarity tasks under both
models, confirming the
Backpack's core property; but differential contextualization recruits far
more of the sense inventory into carrying that signal. The number of
individual senses correlating with human judgments at $\rho>0.6$ rises from
$6/16$ to $9/16$ at $k{=}16$, and from $8/64$ to $37/64$ at
$k{=}64$, with a correspondingly higher mean-senses correlation
($0.58\to0.68$). This is not redundancy: the mean pairwise cosine among a
word's sense vectors is $0.14$, so senses remain nearly orthogonal. The sense lens for \textit{rock} (\S\ref{sec:experiments}, Table~\ref{tab:lens}) shows the same
effect qualitatively: the differential Backpack devotes distinct senses to a word's
separate meanings that the softmax Backpack conflates.

\begin{table}[t]
\centering
\small
\setlength{\tabcolsep}{4pt}
\begin{tabular}{lcccc}
\toprule
Model & best $\rho$ & mean $\rho$ & \#$\rho{>}0.6$ & emb $\rho$ \\
\midrule
softmax, $k{=}16$ & 0.70 & 0.67 & 6/16 & 0.45 \\
diff, $k{=}16$    & 0.73 & 0.69 & 9/16 & 0.42 \\
softmax, $k{=}64$ & 0.72 & 0.58 & 8/64 & 0.48 \\
diff, $k{=}64$    & 0.74 & 0.68 & \textbf{37/64} & 0.48 \\
\bottomrule
\end{tabular}
\caption{Sense specialization on WordSim-353 (Spearman
$\rho$): the single best sense, the sense-averaged cosine, the number of
individual senses with $\rho>0.6$, and the raw token-embedding baseline.}
\label{tab:spec}
\end{table}

\begin{table}[t]
\centering
\small
\begin{tabular}{ll}
\toprule
\multicolumn{2}{c}{\textit{rock}, differential Backpack (ours)} \\
\midrule
mineral & \textit{rock, Rock, Rocks, canyon, Canyon} \\
music   & \textit{rock, garage, punk, gig, pop} \\
\midrule
\multicolumn{2}{c}{\textit{rock}, softmax Backpack} \\
\midrule
blended & \textit{rock, punk, metal, shale} \\
blended & \textit{rock, bedrock, punk, Stones} \\
\bottomrule
\end{tabular}
\caption{Sense lens for \textit{rock}: the vocabulary items each
selected sense most strongly predicts through the tied embedding,
$C(w)_\ell E^\top$, in the two models at $k{=}16$.}
\label{tab:lens}
\end{table}

\section{Robustness Across Sense Counts}
\begin{figure}[t]
  \centering
  \includegraphics[width=\columnwidth]{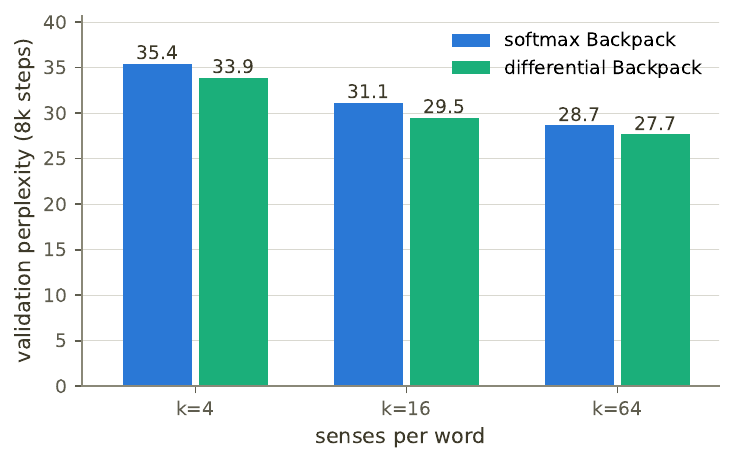}
  \caption{Final perplexity by number of senses $k$ on
  WikiText-103 ($8$k steps).}
  \label{fig:senses}
\end{figure}
The advantage is not tied to a particular sense budget
(Figure~\ref{fig:senses}). Differential contextualization wins at every
sense count we test: $33.3$ vs.\ $34.6$ at $k{=}4$, $28.8$ vs.\ $30.4$
at $k{=}16$, and $27.0$ vs.\ $28.0$ at $k{=}64$ (exact ppl, $8$k steps).
The gain is not monotone in $k$: it peaks at $k{=}16$ ($5.0\%$ relative)
rather than at the smallest inventory ($3.8\%$ at $k{=}4$, $3.4\%$ at
$k{=}64$). The idealized analysis of \S\ref{sec:theory} predicts a
separation that shrinks with $k$, so it does not by itself account for this
profile; at language-model scale the benefit evidently also depends on
having enough senses for inhibition to act on.
\section{Verifying the Separation}
\label{sec:synthetic}
Proposition~\ref{prop:sep} predicts a concrete, measurable floor, so we test
it directly. We construct $m{=}6$ tokens with orthonormal target directions
$u_1..u_m$ and train Backpacks by regression on the combined sense output
$o$ (rather than through the logits: a logit-space probe is confounded,
since a model can win an argmax by inflating embedding norms without leaving
the hull at all; an earlier version of this task was solved by both
architectures for exactly that reason). Each example is a two-token sequence
$[x_i, x_j]$ requiring $o = u_i$ at position $0$ (which pins the senses
exactly) and $o = 1.5\,u_i - 0.5\,u_j$ at position $1$, a point outside
$\mathrm{conv}\{u_i,u_j\}$.

\begin{table}[t]
\centering
\small
\begin{tabular}{lccccc}
\toprule
squared error & $k{=}1$ & $k{=}2$ & $k{=}4$ & $k{=}8$ & $k{=}16$ \\
\midrule
softmax & .275 & .275 & .092 & .028 & .002 \\
diff    & \textbf{.000} & \textbf{.000} & \textbf{.000} & \textbf{.000} & .005 \\
\bottomrule
\end{tabular}
\caption{Extrapolation squared error on the synthetic
task of Appendix~\ref{sec:synthetic} by number of sense heads $k$ (mean of
3 seeds).}
\label{tab:synthetic}
\end{table}

Table~\ref{tab:synthetic} matches the theory. At $k{=}1$ the softmax Backpack
is stuck at a squared error of $0.275$ while the differential Backpack reaches
$0$ to numerical precision, a separation of four orders of magnitude, and
it does so by growing $\lambda$ to $\approx\!0.81$, well past the
$s/(1+s) = 1/3$ threshold the proposition requires. The measured floor is
$0.275$ rather than the proposition's $2s^2 = 0.5$ because the trained
objective also scores the pin, and the model trades a little pinning error
for extrapolation error. This is its true floor, not a poor optimum: with
attention placed entirely on the first token, the joint objective is
minimized at $C(x_i) = 1.25\,u_i - 0.05\sum_{j \ne i} u_j$ (for $m{=}6$,
$s{=}0.5$), which gives pinning error $0.075$ and extrapolation error
$0.275$ exactly, the values observed. At $k{=}2$ the error is unchanged,
but there it is an optimization outcome rather than a floor: a cancelling
solution exists ($C(x)_1 = 1.5\,u_x$, $C(x)_2 = -0.5\,u_x$ with saturated
attention) and training does not find it. From $k{=}4$ the softmax error
decays ($0.092$, $0.028$, $0.002$) as heads cancel, until at $k{=}16$ both
architectures solve the task and the separation disappears.
\section{Inhibition Dynamics}
\label{sec:lambda}
\begin{figure*}[t]
  \centering
  \includegraphics[width=\textwidth]{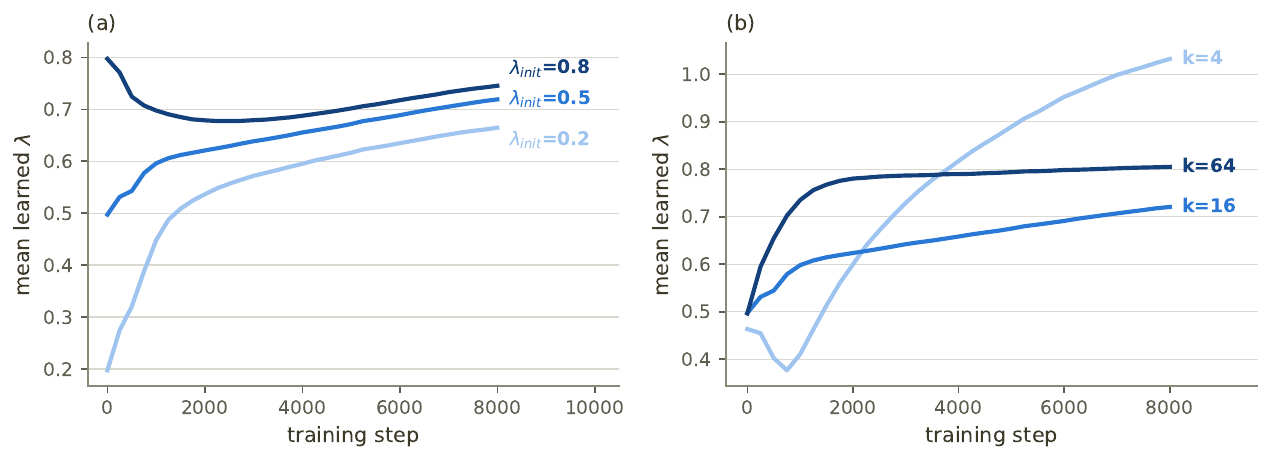}
  \caption{Mean learned $\lambda$ over training (a) from
  three initializations and (b) for three sense counts $k$ (WikiText-103,
  $8$k steps).}
  \label{fig:lambda}
\end{figure*}
How much inhibition does the model want? We track the mean learned gain
$\lambda$ and vary its initialization (Figure~\ref{fig:lambda}). From
initializations $\{0.2, 0.5, 0.8\}$ trained for $8$k steps, $\lambda$
reaches the same band ($0.66$--$0.75$) and final perplexity is nearly flat
across them ($28.8$--$29.4$ at that budget): the level of inhibition is set
by training rather than inherited from the initialization. It is not a
fixed constant, however: in the $20$k-step main run $\lambda$ grows
steadily from $0.5$ to $0.85$ without saturating, so the model relies on
inhibition more, not less, as it improves. Demand for inhibition is highest
when senses are few: mean $\lambda$ is $1.03$ at $k{=}4$, where some senses
become net-inhibitory, against $0.72$ at $k{=}16$ and $0.80$ at $k{=}64$
(all $8$k steps), so it is not monotone in $k$. Across scale at $k{=}16$
the trained gain is stable: $0.85$ at $70$M, $0.84$ at $170$M (three
seeds) and $0.88$ at $350$M.

\paragraph{Is the $\lambda>1$ regime necessary?}
Mean $\lambda$ exceeds $1$ at $k{=}4$, with individual senses reaching
$2.15$, so those senses are net inhibitory and fall outside the regime
Propositions 1 and 2 assume. Constraint \eqref{eq:reach} still holds for
such a sense (its proof never uses $\lambda_\ell<1$), but its normalized
coefficients range over the reflection through the origin of an $\ell_1$
ball section of radius $(1+\lambda_\ell)/(\lambda_\ell-1)$, so the
separation argument of Proposition~\ref{prop:sep}, which compares against
the convex hull, does not apply. Constraining $\lambda$ to $(0,1)$ with a
sigmoid tests whether this regime is necessary or incidental. It costs $0.32$
ppl at $k{=}4$ ($33.30\to33.62$) and nothing measurable at $k{=}16$
($28.84\to28.79$), where trained $\lambda$ stays below $1$ anyway. The
constraint does bind: negative weight mass falls from $0.509$ to $0.411$ at
$k{=}4$, against $0.394$ to $0.381$ at $k{=}16$. The regime is therefore
doing work exactly where it arises: with few, coarse senses, some senses are
most useful as pure subtraction.
\section{Induced versus Learned Senses}
\label{sec:induced}
Because attention-only networks are Backpacks
\citep{hewitt-etal-2023-backpack}, senses can be read out of a trained
Transformer with no training at all: head $\ell$ contributes
$\sum_j \alpha^\ell_{ij}(W_O^\ell W_V^\ell)E(x_j)$, and
$C(x)_\ell := W_O^\ell W_V^\ell E(x)$ depends only on the token. We verified
this reproduces a real attention sublayer to $2\!\times\!10^{-7}$ relative
error. This gives a training-free baseline that learned senses must beat.

\begin{table}[t]
\centering
\small
\setlength{\tabcolsep}{3.5pt}
\begin{tabular}{lccccc}
\toprule
Senses & $k$ & SL999 & SV3.5k & RG65 & WS353 \\
\midrule
induced, Transf. & 112 & 0.322 & 0.179 & 0.493 & 0.475 \\
induced, trunk   & 64  & 0.365 & 0.249 & 0.559 & 0.552 \\
\textbf{learned, diff} & \textbf{16} & \textbf{0.441} & \textbf{0.337} & \textbf{0.726} & \textbf{0.731} \\
\bottomrule
\end{tabular}
\caption{Best-sense Spearman $\rho$ of training-free
senses induced from OV circuits versus learned Backpack senses.}
\label{tab:induced}
\end{table}

Learned senses dominate on all four datasets while using far fewer
components ($4$--$7\times$ fewer; Table~\ref{tab:induced}), so training a Backpack buys a
qualitatively better interface rather than a repackaging of one a
Transformer already has. Consistently, reconstructing the Transformer's
output from its OV circuits alone leaves a relative residual of $0.85$:
most of a Transformer is genuinely outside Backpack form. We report this residual
only for softmax attention; in differential mode the sublayer applies a
per-head RMSNorm, so no $W_OW_V$ factorization exists and the quantity is
undefined.

\section{Word-in-Context: A Negative Result}
\label{sec:wic}
We also tested whether signed sense weights are a better zero-shot
word-sense-disambiguation signal, using WiC
\citep{pilehvar-camacho-collados-2019-wic}: for each sentence pair we take
the cosine similarity between the target word's occurrences under four
representations (the underlying hidden state, the self-position sense
weights $A^{\ell}_{pp}$, the mean downstream sense reads, and the sense
mixture), fit a threshold on train and report dev accuracy. We find
no evidence for the hypothesis. All features reach only
$53$--$58\%$ (majority class $50\%$), and every differential-vs-softmax
difference lies within $1.7$ standard errors ($\pm1.98$ points on $638$
pairs) with the sign flipping between the $70$M and $170$M models. The
contextual hidden-state baseline is itself only $56$--$58\%$, so these
models are too weak at this scale for WiC probing to be diagnostic at all;
the null result bounds our claims rather than testing the mechanism.
One component of the probe is informative, however. Decomposing the signed
self-weights into their inhibitory and excitatory halves and probing each
alone, the inhibitory half reaches $55.3\%$ at $170$M and $54.2\%$ at $70$M,
$2.7$ and $2.1$ standard errors above the $50\%$ majority. For the softmax
Backpack the same feature is identically zero and scores exactly $50.0\%$,
which is the sanity check the decomposition affords. Inhibition alone
therefore carries a small but measurable amount of lexical information
across $638$ pairs, which generalizes the single \textit{bank} analysis of
\S\ref{sec:experiments} beyond one word. We
report the overall null because our disambiguation evidence
(\S\ref{sec:experiments}, Figure~\ref{fig:disambig}) is qualitative, and a
benchmark-level demonstration of improved WSD remains open.

\end{document}